\documentclass[10pt,twocolumn,letterpaper]{article}

\usepackage{cvpr}
\usepackage{times}
\usepackage{epsfig}
\usepackage{graphicx}
\usepackage{amsmath}
\usepackage{amssymb}
\usepackage{multirow}

\usepackage[breaklinks=true,bookmarks=false]{hyperref}

\cvprfinalcopy

\begin{document}

\title{A Simple Probabilistic Model for Uncertainty Estimation}

\author{
  Alexander Kuvaev\\
  {\tt\small a.kuvaev@analysiscenter.ru}
  \and
  Roman Khudorozhkov\\
  {\tt\small r.khudorozhkov@analysiscenter.ru}
}

\maketitle


\begin{abstract}

The article focuses on determining the predictive uncertainty of a model on the example of atrial fibrillation detection problem by a single-lead ECG signal. To this end, the model predicts parameters of the beta distribution over class probabilities instead of these probabilities themselves.

It was shown that the described approach allows to detect atypical recordings and significantly improve the quality of the algorithm on confident predictions.

\end{abstract}


\section{Introduction}

Despite the fact that deep learning architectures have been successfully applied to a large variety of supervised learning tasks, they tend to be overconfident in their predictions even when inference is performed on unusual examples, drawn from the distribution, different from that of the training set. If the cost of a mistake is high, for example in the case of medical diagnosis problem, an additional mechanism of predictive uncertainty estimation is required to avoid making a potentially wrong decision. Unfortunately, classical models lack this ability. However, several ways to address this problem exist, the majority of them are somehow connected with the Bayesian approach to machine learning.

Model's uncertainty in the Bayesian framework can be obtained by estimating the chosen measure of the ``spread'' of the posterior predictive distribution. But inference in Bayesian neural networks is difficult due to the fact that this distribution can not be obtained in a closed form for complex models and approximate methods have to be used, such as Monte Carlo integration by mixing predictive distributions for different sets of weights, drawn from their posterior distribution. Moreover, compared to non-Bayesian neural networks, they are harder to implement, require significant modifications to the training procedure and are computationally more expensive.

The article considers a simple way of model uncertainty estimation by directly learning parameters of the predictive distribution with an ordinary neural network based on the example of atrial fibrillation detection problem.

The proposed model is implemented in CardIO framework~\cite{cardio_2017_1156085} available at \url{https://github.com/analysiscenter/cardio}.


\section{Atrial fibrillation}

Atrial fibrillation (also called AF or AFib) is the most common heart arrhythmia, occurring in about 2\% of the world's population. It is associated with significant mortality and morbidity from heart failure, dementia and stroke. The early AF identification is an essential part of preventing the development of heart diseases, but it is a challenging task due to its episodic nature and similarity to many other abnormal rhythms.


\section{Dataset}

The publicly available training part of the 2017 PhysioNet/CinC Challenge dataset~\cite{PhysioNet2017, PhysioNet} was used for model training and testing. It is a set of 8,528 single-lead ECG recordings lasting from 9 to 61 seconds with an equal sampling rate of 300 Hz. All ECGs were collected from portable heart monitoring devices and manually classified by a team of experts into 4 classes: atrial fibrillation, normal rhythm, other rhythm or too noisy to be classified. A significant part of the signals had their R-peaks directed downwards since the device did not require the user to hold it in any specific orientation. We did not use noisy signals and focused on solving a two-class classification problem: atrial fibrillation against normal and other rhythms. The dataset was randomly split into 80\% training and 20\% validation subsets.


\section{Model}


\subsection{Architecture}

For this learning task, a convolutional neural network was used. However, instead of predicting class probabilities themselves, the network predicts parameters of the Dirichlet distribution over these probabilities. Since a two-class classification task is considered, the Dirichlet distribution comes down to its special case in the form of the beta distribution. It was chosen here instead of the more common Bernoulli distribution because the latter is unable to represent a certain prediction significantly different from 0 and 1 in sense of variance or entropy of the predictive distribution. This fact will be discussed in more detail later in section~\ref{discussion}.

The network architecture is as follows. First, cropped segments are passed through a convolutional layer with 8 filters with a kernel size of 5 followed by a max pooling operation with a window size and a stride of 2 and a ReLU activation~\cite{relu}.

Next comes a sequence of classical ResNet blocks~\cite{resnet}, combined into groups. Blocks in each group have the same parameters, shown in Table~\ref{tab:architecture}. The first block in each group performs a downsampling operation along the spatial dimension with a stride of 2. ReLU activation is used in all residual units.

The network ends with a global max pooling operation and a fully-connected layer with a softplus activation to return positive parameters of the beta distribution.

The convolution stride in all layers, except the first layer in each group of ResNet blocks, is fixed to 1. The input of each convolutional layer is padded in such a way that the spatial resolution is preserved afterwards. Batch normalization~\cite{BN} is used before each activation. All convolutional kernels were initialized by the scheme, proposed in~\cite{xavier}, the biases were initialized with zeros.

Adam optimizer~\cite{adam} with a mini-batch size of 256 signals was used for model training. The exact network configuration is shown in Table~\ref{tab:architecture}.

\begin{table}[t]
\begin{center}
\def\arraystretch{1.1}
\begin{tabular}{c | c | c}
\hline
Layer        & Output size & Block parameters                                \\ \hline
Input        & 2048        &                                                 \\ \hline
Conv block   & 1024        & 5, 8, /2                                        \\ \hline
ResNet block & 512         & \( \bigg[ \hspace{-0.4em} \begin{array}{l}
                                3, \; 8 \\
                                3, \; 8
                              \end{array} \hspace{-0.4em} \bigg] \times 2 \) \\ \hline
ResNet block & 256         & \( \bigg[ \hspace{-0.4em} \begin{array}{l}
                                3, \; 8 \\
                                3, \; 8
                              \end{array} \hspace{-0.4em} \bigg] \times 2 \) \\ \hline
ResNet block & 128         & \( \bigg[ \hspace{-0.4em} \begin{array}{l}
                                3, \; 12 \\
                                3, \; 12
                              \end{array} \hspace{-0.4em} \bigg] \times 2 \) \\ \hline
ResNet block & 64          & \( \bigg[ \hspace{-0.4em} \begin{array}{l}
                                3, \; 12 \\
                                3, \; 12
                              \end{array} \hspace{-0.4em} \bigg] \times 2 \) \\ \hline
ResNet block & 32         & \( \bigg[ \hspace{-0.4em} \begin{array}{l}
                                3, \; 16 \\
                                3, \; 16
                              \end{array} \hspace{-0.4em} \bigg] \times 3 \) \\ \hline
ResNet block & 16          & \( \bigg[ \hspace{-0.4em} \begin{array}{l}
                                3, \; 16 \\
                                3, \; 16
                              \end{array} \hspace{-0.4em} \bigg] \times 3 \) \\ \hline
ResNet block & 8           & \( \bigg[ \hspace{-0.4em} \begin{array}{l}
                                3, \; 20 \\
                                3, \; 20
                              \end{array} \hspace{-0.4em} \bigg] \times 2 \) \\ \hline
\multirow{2}{*}{\shortstack{Classification\\ layer}} & 1 & global max pooling      \\ \cline{2-3} 
                                                     &   & fully-connected, softplus \\ \hline
\end{tabular}
\end{center}
\caption{Proposed network architecture.}
\label{tab:architecture}
\end{table}


\subsection{Training}

The model was trained end-to-end on approximately 7 second long segments from ECG signals by minimizing the negative beta log-likelihood. Under the assumption that heart rhythm does not change along the whole length of the recording, each segment was labeled with the class of the original signal.

All signals whose R-peaks were directed downwards were flipped before cropping. The number of crops for each signal depended on the signal's label in order to balance the classes in each batch. Also, signal resampling to a new randomly selected sampling rate was used as a data augmentation step, thus allowing the model to adapt to a broader range of possible heartbeat rates.


\subsection{Inference}

Making a prediction in such a model is not so straightforward. A new ECG signal can have an arbitrary length, while the network is trained on fixed-sized crops. Therefore, an algorithm for multiple predictions aggregation is needed.

Let's denote the signal generating process by \(X,\) the atrial fibrillation probability by \(t\) and the vector of beta distribution parameters, obtained deterministically from the network, by \(\alpha.\) Consider the conditional distribution of \(t\) given \(X:\)
\begin{align*}
	p(t \,|\, X) &= \int p(t \,|\, \alpha, X) p(\alpha \,|\, X) \; d\alpha = \\
    &=\mathbb{E}_{p(\alpha \,|\, X)} \text{Beta}(t \,|\, \alpha) \approx \frac{1}{N} \sum_{i = 1}^{N} \text{Beta}(t \,|\, \alpha_{i}),
\end{align*}
where \( \alpha_{i} \thicksim p(\alpha \,|\, X). \) Suppose \(X\) is an ergodic process, which is a reasonable assumption if the beat type remains unchanged during the whole recording. In this case, samples from \(p(\alpha \,|\, X)\) may be replaced by outputs of the network for consequent non-overlapping crops from the original signal.

Thus, the distribution over atrial fibrillation probability can be approximately modeled by the mixture of beta distributions with equal weights. The mean of the mixture provides a point estimate of this probability. The predictive uncertainty can be defined as the variance of the mixture multiplied by 4. This multiplier comes from the fact that the variance of a random variable, whose support is a subset of the [0, 1] interval, is bounded both below and above by 0 and 0.25 respectively, therefore the uncertainty measure, defined in this way, takes values from 0 (absolutely sure) to 1 (absolutely unsure).


\section{Results}

\begin{figure*}
\begin{center}
\includegraphics[width=\linewidth]{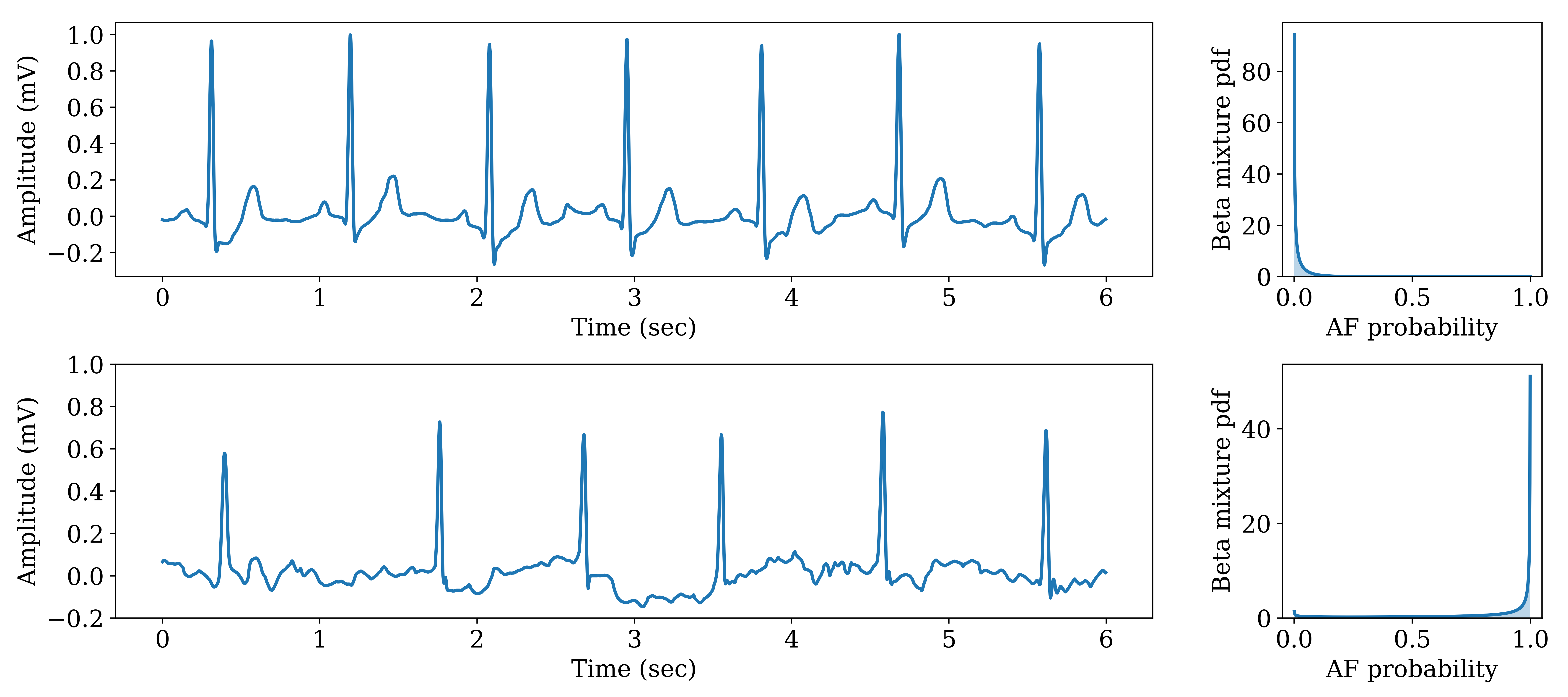}
\end{center}
\vspace{-12pt}
\caption{Examples of certain predictions. Top: ECG with normal rhythm. Bottom: ECG with atrial fibrillation.}
\label{fig:certain}
\end{figure*}

\begin{figure*}
\begin{center}
\includegraphics[width=\linewidth]{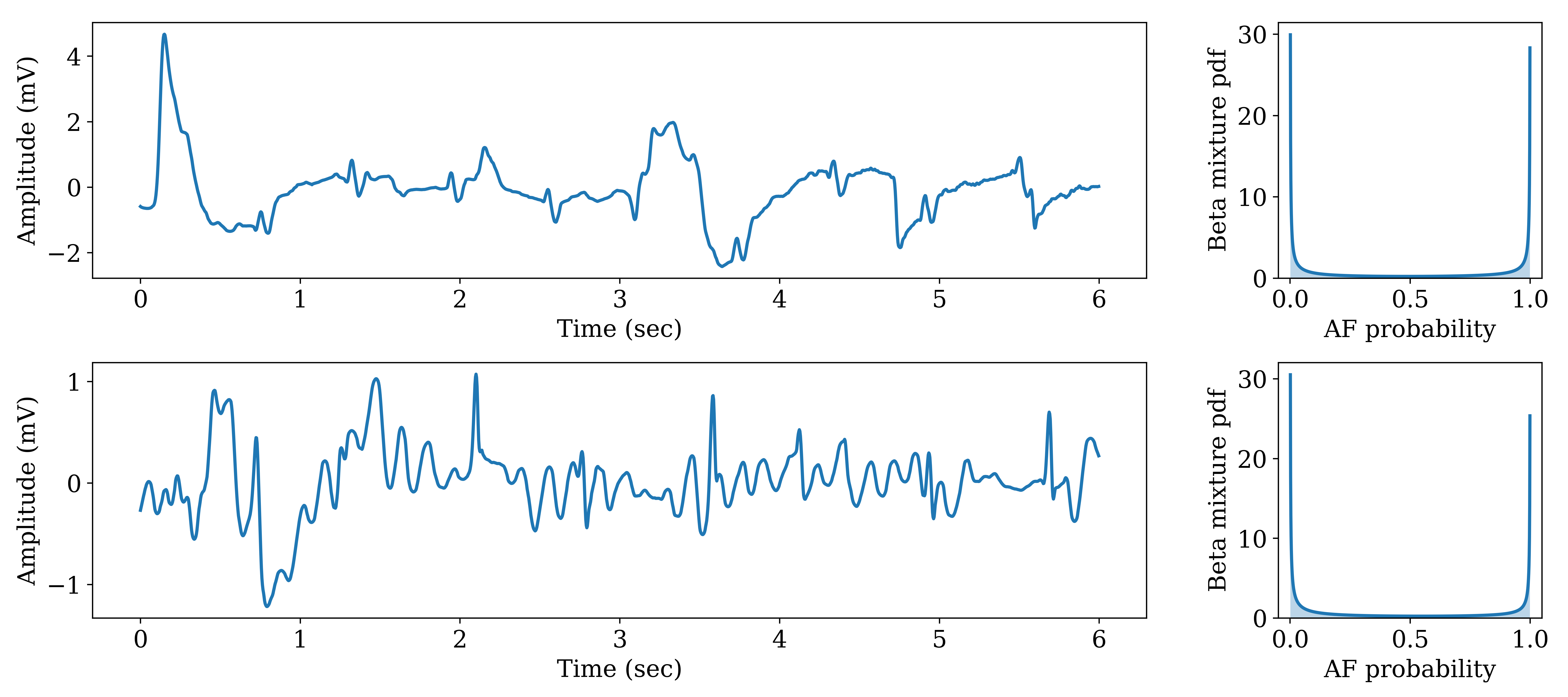}
\end{center}
\vspace{-12pt}
\caption{Examples of uncertain predictions.}
\label{fig:uncertain}
\end{figure*}

The defined uncertainty measure allows us to select a threshold separating certain predictions from uncertain ones, for which the classification will be denied. Tables~\ref{tab:all_metrics} and~\ref{tab:certain_metrics} show precision, recall and F1-score for both atrial fibrillation (``A'') and normal and other rhythm (``NO'') classes as well as macro-averaged values of these metrics for the whole validation set and 90\% most certain predictions respectively. We can observe a significant increase in overall F1-score caused by improved classification performance for the atrial fibrillation class. Also, after the removal of uncertain predictions, the number of misclassified ECGs decreased from 58 to only 16 signals.

\begin{figure*}
\begin{center}
\includegraphics[width=\linewidth]{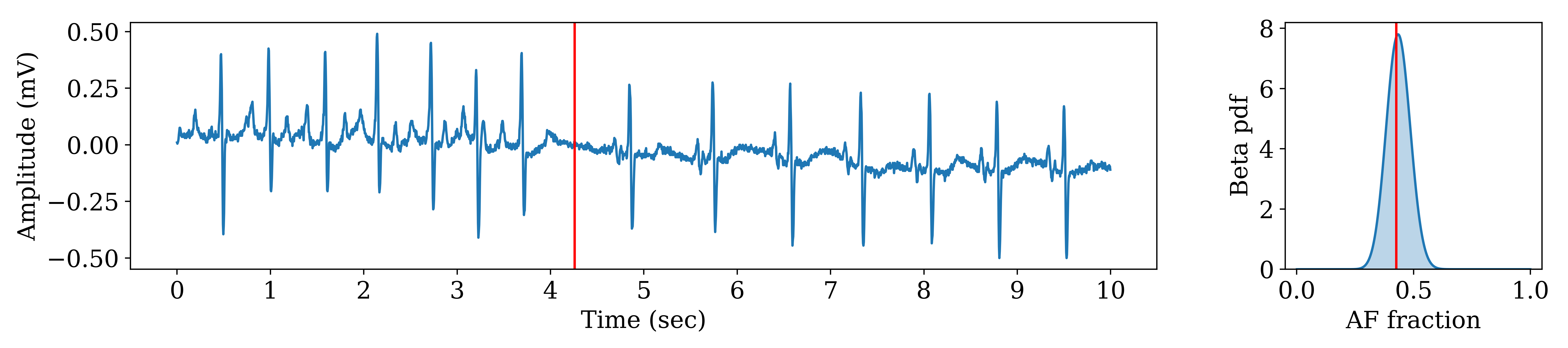}
\end{center}
\vspace{-12pt}
\caption{Left: Sampled segment from an ECG recording. The red line represents the boundary of a heartbeat change. Right: The pdf of the predictive distribution over a fraction of an arrhythmic part in the entire segment. The red line represents the actual fraction.}
\label{fig:mit-bih}
\end{figure*}

Now let’s look at two examples of certain predictions shown in figure~\ref{fig:certain}. The left plot on the top illustrates a healthy person’s ECG, that has a clear quasi-periodic structure. The bottom-left plot, by contrast, demonstrates an ECG with atrial fibrillation: it has irregular R-R intervals with characteristic waves between them. The right plots show the pdf of the predictive distributions with atrial fibrillation probability plotted on the horizontal axis. In both these cases, the model is correct and absolutely sure about its decision: the variance of the distribution is almost 0. Two examples of uncertain predictions are shown in figure~\ref{fig:uncertain}. These are ECGs with irregular structure, which may be caused by a disease or some measurement errors. The probability density on the right plots is almost equally concentrated around 0 and 1 achieving almost maximal possible variance.

\begin{table}[t]
\begin{center}
\def\arraystretch{1.1}
\begin{tabular}{l|c|c|c}
\hline
Class   & Precision & Recall & F1-score \\ \hline
A       & 0.85      & 0.81   & 0.83     \\ \hline
NO      & 0.98      & 0.98   & 0.98     \\ \hline
Overall & 0.91      & 0.89   & 0.90     \\ \hline
\end{tabular}
\end{center}
\caption{Precision, recall and F1-score for atrial fibrillation class (``A''), normal and other rhythm class (``NO'') and overall macro-averaged values of these metrics for the whole validation set.}
\label{tab:all_metrics}
\end{table}

\begin{table}[t]
\begin{center}
\def\arraystretch{1.1}
\begin{tabular}{l|c|c|c}
\hline
Class   & Precision & Recall & F1-score \\ \hline
A       & 0.95      & 0.89   & 0.92     \\ \hline
NO      & 0.99      & 1.00   & 0.99     \\ \hline
Overall & 0.97      & 0.95   & 0.96     \\ \hline
\end{tabular}
\end{center}
\caption{Precision, recall and F1-score for atrial fibrillation class (``A''), normal and other rhythm class (``NO'') and overall macro-averaged values of these metrics for 90\% most certain predictions in the validation set.}
\label{tab:certain_metrics}
\end{table}

All this means that the chosen uncertainty measure actually reflects model's uncertainty in its prediction and allows to detect atypical recordings by comparing the uncertainty with a predefined threshold.


\section{Discussion} \label{discussion}

As can be seen from figures~\ref{fig:certain} and~\ref{fig:uncertain}, the density of the predictive distribution tends to concentrate around 0, 1 or both these values. This is indeed true for the majority of signals in the validation set and is due to minimization of negative log-likelihood of hard labels during training.

In case of hard labels, the behavior of the model is very similar to that of the network with the same architecture, where the last layer just predicts atrial fibrillation probability directly with a sigmoid activation. However, in case of soft labels, such a model is unable to represent a certain prediction significantly different from 0 and 1 in sense of variance or entropy of the predictive distribution, while the described model can.

Soft labels for an atrial fibrillation detection problem can be obtained, for example, from the MIT-BIH Atrial Fibrillation Database~\cite{MIT-BIH}. The dataset includes 25 long-term ECG recordings along with points of heartbeat change for each signal. Such annotation allows to sample segments around these points and define the target as a fraction of an arrhythmic part in the entire segment. The proposed model was trained on this dataset, an example of a certain prediction is shown in figure~\ref{fig:mit-bih}. It can be seen that the predictive pdf, produced by the model, has a bell-shaped form rather than being concentrated at the ends of the [0, 1] interval.

Another way of obtaining soft labels is to use methods transforming hard labels into soft ones, such as mixup~\cite{mixup}.


\section{Conclusion}

On the example of atrial fibrillation detection problem by a single-lead ECG signal, a simple approach for predictive uncertainty estimation is described. The proposed model predicts parameters of the beta distribution over class probabilities instead of these probabilities themselves. This approach allows to detect atypical recordings by comparing the uncertainty with a predefined threshold and significantly improve the quality of the algorithm on confident predictions.


{\small
\bibliographystyle{ieee}
\bibliography{egbib}
}

\end{document}